\documentclass[10pt,twocolumn,letterpaper]{article}

\usepackage{wacv}              % To produce the CAMERA-READY version
\usepackage{cuted}   % 关键包，用于强制单栏区域
\definecolor{wacvblue}{rgb}{0.21,0.49,0.74}
\usepackage[pagebackref,breaklinks,colorlinks,allcolors=wacvblue]{hyperref}
\usepackage{booktabs,multirow,pifont,tabularx,makecell}
\newcommand{\cmark}{\ding{51}} % ✓
\newcommand{\xmark}{\ding{55}} % ✗
\usepackage{multirow}
\usepackage[table]{xcolor}
\usepackage{booktabs}
\newcommand{\TMImean}{\ensuremath{\mathrm{TMI}_{\mathrm{mean}}}}
\newcommand{\TMImed}{\ensuremath{\mathrm{TMI}_{\mathrm{med}}}}
\newcommand{\TMinImean}{\ensuremath{\mathrm{TMinI}_{\mathrm{mean}}}}
\newcommand{\TIoU}{\ensuremath{\mathrm{TIoU@0.5}}}
\newcommand{\TCov}{\ensuremath{\mathrm{TCov}}}

\def\wacvPaperID{16} % *** Enter the WACV Paper ID here
\def\confName{WACV}
\def\confYear{2027}

\title{Towards Accurate and Robust Surveillance Roadside IVD \\ via Trackletized Audio-Visual Reasoning}

\author{%
\makebox[\textwidth][c]{%
\begin{tabular}{@{}p{0.42\textwidth}@{\hspace{0.06\textwidth}}p{0.42\textwidth}@{}}
\centering Xiwen Li$^{1,2}$ &
\centering Xiaoya Tang$^{1,2}$ \tabularnewline
\centering {\tt\small xiwen.li@utah.edu} &
\centering {\tt\small xiaoya.tang@utah.edu} \tabularnewline[0.65em]
\centering Bodong Zhang$^{2,3}$ &
\centering Tolga Tasdizen$^{2,3}$ \tabularnewline
\centering {\tt\small Bodong.Zhang@utah.edu} &
\centering {\tt\small tolga.tasdizen@utah.edu}
\end{tabular}
}\\[0.7em]
$^1$Kahlert School of Computing, University of Utah\\
$^2$Scientific Computing and Imaging Institute, University of Utah\\
$^3$Department of Electrical and Computer Engineering, University of Utah\\
Salt Lake City, USA
}

\begin{document}
\maketitle
% % ---- FORCE SINGLE COLUMN FIGURE BEFORE ABSTRACT ----
% \begin{strip}
%     \centering
%     \fbox{\rule{0pt}{0.25\textheight} \rule{\textwidth}{0pt}}
%     % \includegraphics[width=\textwidth, height=0.25\textheight, keepaspectratio]{example-image} % 换成你的图片
%     \captionof{figure}{This is a placeholder figure inserted between title and abstract.}
%     \label{fig:example}
% \end{strip}
\begin{abstract}
Idling Vehicle Detection (IVD) seeks to determine, at the final frame of a video clip, whether any vehicle is idling, meaning the vehicle is stationary with its engine running, using synchronized video from a remote surveillance camera and multichannel audio captured by spatially distributed wireless microphones along the roadside \cite{Li2023RealTimeIV, Li2024JointAI, Li2025HAVTIVDHC}. Prior full-image, clip-level fusion approaches \cite{Li2024JointAI, Li2025HAVTIVDHC} tend to overfit scene background and full-frame context, produce unstable temporal decisions, and lack an explicit spatial prior to align vehicles with microphones, which makes them brittle under domain shift and data inefficient. Instead, we introduce TAVR-IVD, an audio-visual framework guided by multi-object tracking. Our method detects vehicles, links detections into tracklets, and classifies each vehicle by operating on its tracklet. This design raises the effective signal-to-noise ratio, stabilizes temporal decisions through tracklets, enforces an explicit spatial prior to align vehicles with microphones, and adapts across domains with limited calibration annotations while remaining detector agnostic and efficient. To evaluate deployment robustness, we further curate two evaluation extensions, AVIVD-LT and AVIVD-M, covering inter-day and cross-site shifts.
\end{abstract}
    
\section{Introduction}
\label{sec:intro}
% p1
An idling vehicle---defined as a stationary vehicle with its engine running---contributes to air pollution and unnecessary fuel consumption. Leveraging remote microphones and a webcam in a roadside surveillance setup offers a practical and scalable solution for monitoring vehicle idling behavior, enabling Idling Vehicle Detection (IVD) to be formulated as an audio-visual detection problem. Formally, given a video clip $V$ and synchronized multi-channel audio signals, the task of IVD is to predict each vehicle's status (moving, idling, or engine-off) at the last frame. Unlike many audio-visual tasks, IVD requires modeling three distinct forms of cross-modal correspondence: (i) \emph{moving}---vehicle motion accompanied by relatively high-RPM engine sound; (ii) \emph{idling}---a stationary vehicle with low-RPM engine sound; and (iii) \emph{engine-off}---a stationary vehicle with no engine sound. Effectively learning these correspondence patterns is therefore essential for accurate IVD.

% p2
Existing end-to-end IVD methods (e.g., AVIVDNet \cite{Li2024JointAI} and HAVT-IVD \cite{Li2025HAVTIVDHC}) typically perform clip-level audio-visual fusion by applying 3D CNN backbones to the entire video clip with full spatial frames. While effective under the original evaluation setting, this formulation couples the decision to global scene context and often entangles target vehicles with background cues (road layout, illumination, static structures, and ambient traffic), making the learned model prone to shortcut learning and thus fragile to distribution shifts. More fundamentally, roadside IVD exhibits a global-to-instance mismatch: multi-channel audio is captured as a global mixture of multiple sound sources, whereas the desired output is a per-vehicle status at the last frame. In practice, building an accurate and deployable IVD system requires four key capabilities:
(1) \textbf{Instance Anchoring} to ground predictions on individual vehicles rather than the whole scene;
(2) \textbf{Geometry-conditioned Binding} to associate globally mixed multi-channel audio with the correct vehicle instance under sensor/scene geometry;
(3) \textbf{Temporal Consolidation} to align temporally distributed, noisy cues with the last-frame labeling protocol and avoid unstable frame-level decisions; and
(4) \textbf{Domain-robust Representation Shaping} to suppress domain-specific shortcuts and generalize across deployment days and locations.
However, prior work is mainly evaluated under same-day and same-scene settings, leaving \emph{inter-day} and \emph{cross-site} generalization largely untested. As a result, existing E2E IVD models can appear strong in-domain but may fail to reliably generalize to data collected on a different deployment day or at a different deployment site. To bridge this gap, we introduce explicit inter-day and cross-site test sets, and systematically evaluate generalization under both types of deployment shifts.

% p3
To address these challenges, we propose \textbf{Trackletized Audio-Visual Reasoning (TAVR)}, a tracklet-centric framework that reformulates IVD from full-frame clip classification into per-tracklet cross-modal reasoning. By operating on compact, instance-centric tracklets rather than full spatial frames, TAVR reduces background entanglement and simplifies the visual input, improving both accuracy and generalization. Concretely, TAVR first performs \textbf{instance anchoring} by converting frame-level detections into identity-consistent vehicle tracklets; importantly, we leverage a strong off-the-shelf detector (e.g., YOLO) and, when needed, calibrate it with only a small amount of target-domain \emph{bounding-box} annotations to rapidly obtain reliable tracklets in new deployment domains. Built on these anchors, \textbf{MASP} enables \textbf{geometry-conditioned binding} by conditioning audio--visual association on vehicle geometry, facilitating instance-specific attribution from globally mixed multi-channel audio. Next, a \textbf{tracklet-conditioned classifier} performs \textbf{temporal consolidation} by aggregating multimodal evidence along each tracklet to produce stable per-vehicle status predictions. Finally, \textbf{JACE} enforces a status-structured cross-modal latent space, providing \textbf{domain-robust representation shaping} that mitigates shortcut learning and improves robustness under deployment shifts. Overall, TAVR moves IVD \emph{towards accurate and generalizable} roadside surveillance under realistic \emph{inter-day} and \emph{cross-site} deployment shifts.

% p4: main contributions
We summarize our contributions in the following four-fold manner:
\begin{itemize}
    \item \textbf{TAVR Framework.} We propose \textbf{Trackletized Audio-Visual Reasoning (TAVR)} for roadside IVD, which localizes vehicles per frame, constructs clip-wise vehicle tracklets via multi-object tracking, and performs per-vehicle status inference from multi-channel audio and tracklets.
    \item \textbf{Geometry-conditioned binding and domain-robust learning.} We introduce \textbf{MASP} for geometry-conditioned audio--tracklet binding and \textbf{JACE} for status-structured cross-modal representation shaping, improving robustness against shortcut learning and deployment shifts.
    \item \textbf{New evaluation datasets for generalization.} To enable rigorous evaluation beyond same-day, same-scene settings, we curate two new test datasets: \textbf{AVIVD-LT} for \emph{inter-day} generalization, and \textbf{AVIVD-M} for \emph{cross-site} generalization.
    \item \textbf{Accuracy and generalizability.} Our approach achieves SOTA performance on the AVIVD validation set and the newly constructed test sets, demonstrating both high accuracy and improved robustness under domain shifts compared to prior end-to-end baselines.
\end{itemize}

\begin{figure*}[htbp]
    \centering
    \includegraphics[width=0.92\linewidth]{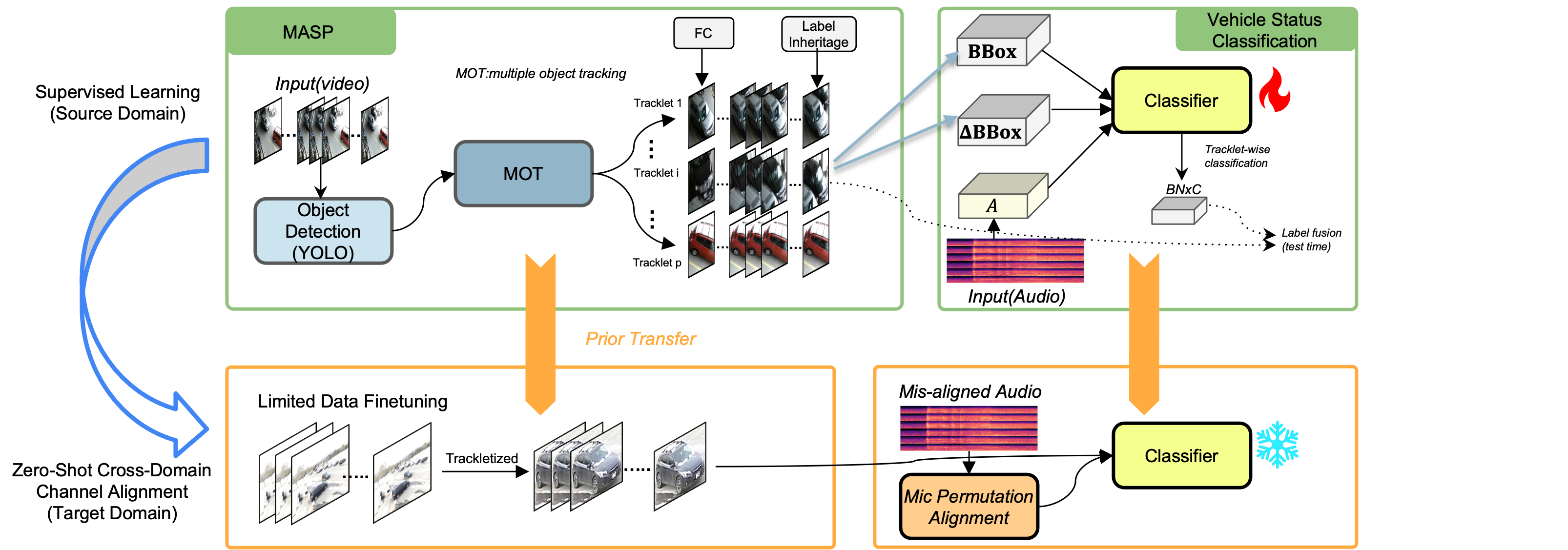} % 换成你的图片
    \caption{TAVR Framework with Training-Free Cross-Domain Channel Alignment.}
    \label{fig:TAVR}
\end{figure*}
%-------------------------------------------------------------------------

\section{Related Work}
\noindent\textbf{Idling Vehicle Detection.} Idling Vehicle Detection (IVD) has evolved from early sensor-specific formulations to recent audio-visual learning-based approaches. Early work explored infrared imagery and deep spatio-temporal modeling to distinguish idling vehicles from stopped vehicles based on thermal signatures, demonstrating the feasibility of automatic IVD under specialized sensing setups \cite{bastan2020remote}. Later, audio-visual methods introduced more practical roadside surveillance solutions by combining remote microphones and RGB video. A representative real-time system adopted a disjoint pipeline that first detects vehicle motion visually and then uses a contrastive latent space to classify stationary vehicles as idling or engine-off from audio cues \cite{Li2023RealTimeIV}. More recent work shifted toward end-to-end joint audio-visual detection: AVIVDNet \cite{Li2024JointAI} performs clip-level fusion with streamlined input dependencies for fully automatic deployment, while HAVT-IVD \cite{Li2025HAVTIVDHC} further improves cross-modal alignment with transformer-based fusion, multiscale visual features, and decoupled detection heads. In contrast to these methods, which mainly rely on full-frame clip-level fusion and implicitly entangle target vehicles with scene context, our work reformulates IVD as \emph{tracklet-centric} audio-visual reasoning, explicitly grounding prediction on individual vehicle instances and enabling more accurate and generalizable deployment.

\noindent\textbf{Audio-Visual Localization, Tracking, Active Speaker Detection, and Generic Models.}
Beyond IVD, audio-visual learning has been studied for vehicle localization, object tracking, sound source localization, active speaker detection, and generic multimodal representation learning.
Audio-visual vehicle localization and tracking methods show that sound can provide useful cues for detecting or tracking vehicles, often through cross-modal distillation from visual teachers to audio students~\cite{gan2019self, huang3, huang2024ar, huang_ai-augmented_2025, tang2022few, liu2023spts, MM-Distill}.
Audio-visual sound source localization and segmentation methods, such as VGG-SS~\cite{Chen2021LocalizingVS} and AVSBench~\cite{zhou2022avs}, further learn to associate audio with sounding regions or pixels in visual scenes.
Another closely related line is active speaker detection (ASD), where AVA-ActiveSpeaker provides face-track-level speaking annotations~\cite{roth2020ava}, and subsequent methods such as TalkNet~\cite{tao2021talknet}, MAAS~\cite{alcazar2021maas}, Light-ASD~\cite{liao2023lightasd}, and LoCoNet~\cite{wang2024loconet} perform instance-level audio-visual reasoning over face tracks or lip/face crops.
These ASD methods demonstrate the value of localizing an instance before audio-visual classification, but they rely on strong visual motion cues from the mouth region.
In contrast, roadside IVD requires reasoning over vehicle tracklets, where idling and engine-off vehicles can be visually static and nearly indistinguishable from cropped appearance alone.
Recent generic audio-visual representation models, including ImageBind~\cite{girdhar2023imagebind}, LanguageBind~\cite{zhu2023languagebind}, AV-HuBERT~\cite{shi2022learning}, and CAV-MAE~\cite{gong2023contrastive}, as well as audio-video MLLMs such as MiniCPM-o~\cite{minicpmo2025} and VideoLLaMA 2~\cite{damonlpsg2024videollama2}, provide transferable audio-video understanding ability.
However, these models are not designed for per-vehicle engine-state reasoning under globally mixed multi-channel roadside audio.
Our work therefore differs from prior audio-visual localization, ASD, and generic multimodal models by using vehicle tracklets as instance anchors and compact tracklet geometry as a spatial prior for multi-channel audio attribution and final-frame status classification.

\noindent\textbf{Contrastive Representation Learning.} Contrastive learning has been widely studied for learning discriminative and transferable representations. 
Early metric-learning work introduced contrastive loss to pull similar samples together and push dissimilar samples apart~\cite{hadsell2006contrastive}. 
More recent self-supervised methods, including CPC~\cite{oord2018cpc}, instance discrimination~\cite{wu2018instance}, MoCo~\cite{he2020moco}, and SimCLR~\cite{chen2020simclr}, show that contrastive objectives can learn representations that transfer well to downstream recognition, detection, and low-label settings. 
In multimodal learning, CMC~\cite{tian2020cmc} and CLIP~\cite{radford2021clip} further demonstrate that contrastive learning can align different views or modalities into a shared representation space, enabling strong transfer and zero-shot generalization. 
Supervised contrastive learning (SupCon)~\cite{khosla2020supcon} extends this idea to labeled data by treating samples from the same class as positives and samples from different classes as negatives, leading to more compact and class-discriminative embeddings. 
Motivated by these works, our JACE module applies supervised contrastive learning to joint tracklet-geometry and global multi-channel audio features. 
Unlike generic contrastive pretraining, JACE defines positives and negatives using vehicle status labels and constructs a task-specific audio-geometry latent space for moving, idling, and engine-off vehicles, improving cross-domain generalization for roadside IVD.

\section{Proposed Method}

\subsection{Problem Setup and Evaluation Splits} 
Given a video clip $V$ and synchronized multi-channel audio $A$, the goal of IVD is to predict the status of each vehicle visible in the final frame. Each vehicle is assigned one of three labels: moving, idling, or engine-off. We denote the original AVIVD \cite{Li2024JointAI} validation split as AVIVD-LV. To evaluate deployment generalization beyond the original same-day setting, we additionally construct two test-only extensions: AVIVD-LT, collected at the same location on a different day, and AVIVD-M, collected at a different deployment site with different microphone configurations. AVIVD-LT evaluates inter-day generalization, while AVIVD-M evaluates cross-location and sensor-layout generalization. Unless otherwise stated, TAVR is trained on the AVIVD training split, and no vehicle-status labels from AVIVD-LT or AVIVD-M are used to train the status classifier or JACE. For AVIVD-M, target-domain labels are used only for the training-free channel-correspondence selection described in Sec.~\ref{sec:channel_alignment}; model parameters remain unchanged during this step.

\subsection{TAVR Framework.} \label{sec:MOTG-IVD} 
Roadside IVD~\cite{Li2023RealTimeIV, Li2024JointAI, Li2025HAVTIVDHC} is inherently an object-level inference problem under a mixed-sensor observation setting. 
Multi-channel audio is recorded globally as a mixture of multiple sound sources, while the desired output is the status of each vehicle at the final frame. 
This global-audio to instance-output mismatch makes full-frame audio-visual fusion prone to background shortcut learning and unstable predictions, especially in multi-vehicle scenes. 

\begin{figure}[htbp]
    \centering
    \includegraphics[width=\columnwidth]{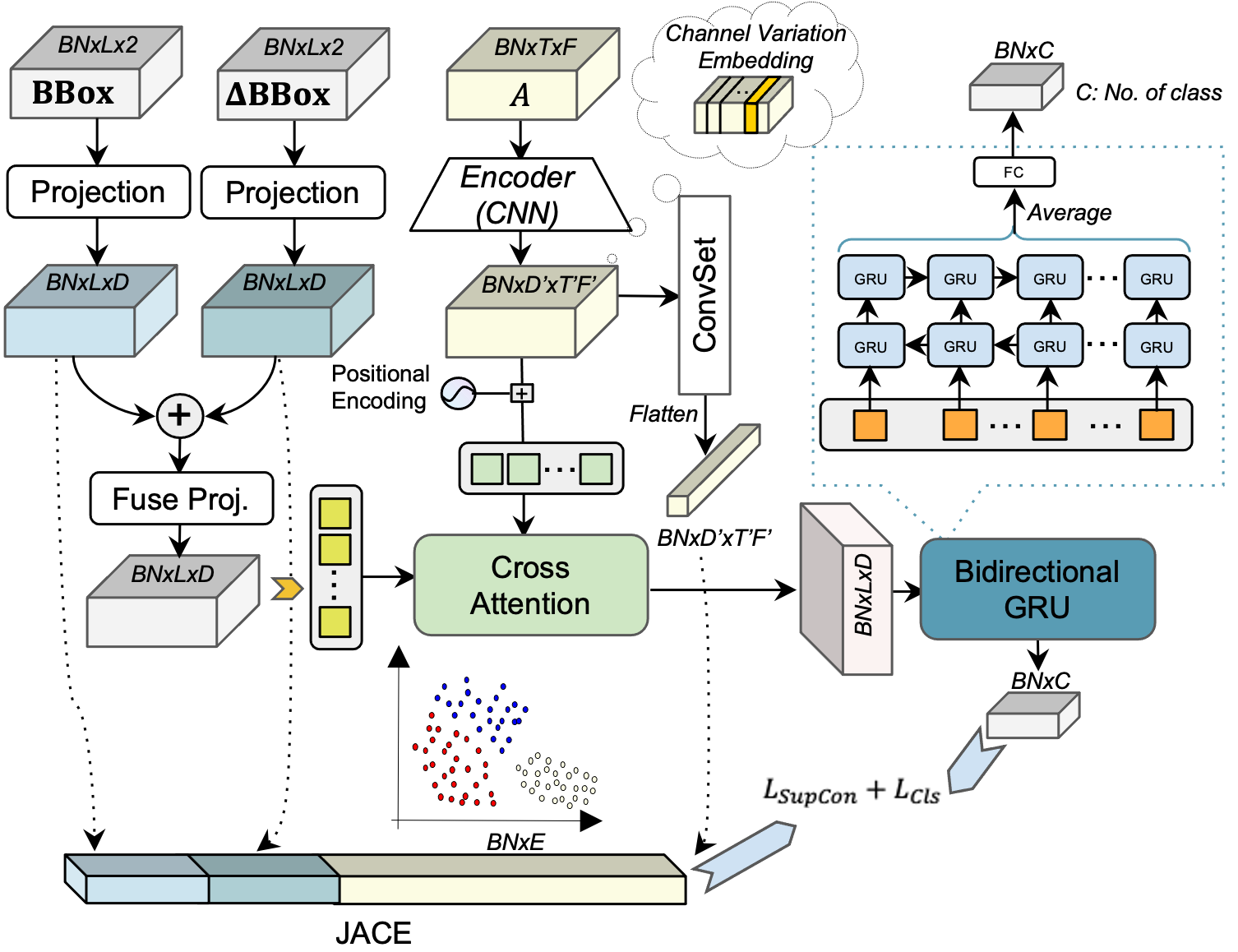}
    \caption{TAVR Classifier.}
    \label{fig:tavr_classifier}
\end{figure}
Prior end-to-end IVD models such as AVIVDNet~\cite{Li2024JointAI} and HAVT-IVD~\cite{Li2025HAVTIVDHC} operate on full surveillance frames. 
Although effective in-domain, this formulation exposes the model to large amounts of irrelevant scene context and can entangle vehicle-status prediction with domain-specific cues such as road layout, illumination, static structures, and surrounding traffic. 
Moreover, jointly learning localization and subtle engine-state classification in a single full-frame detector can lead to brittle optimization. 

We therefore reformulate IVD as trackletized audio-visual reasoning (TAVR). 
As shown in \cref{fig:TAVR}, TAVR first constructs identity-consistent vehicle tracklets with MASP, then performs tracklet-conditioned status classification by combining multi-channel audio, centroid position, and centroid displacement, and finally uses JACE to organize the joint audio-geometry embedding into a status-structured latent space for improved cross-domain robustness.

\noindent\textbf{Mic-Aligned Spatial Prior (MASP).}
As shown in \cref{fig:TAVR}, TAVR uses vehicle tracklets as instance anchors for audio-geometry reasoning.
We first employ YOLO11~\cite{ultralytics_yolo11_2024} to produce frame-wise vehicle bounding boxes within the target roadside region.
Since IVD only considers vehicles inside the predefined monitoring area, we fine-tune YOLO11 with bounding-box annotations and do not use vehicle-status labels during detector training.
After frame-wise detection, we apply DeepSORT~\cite{wojke2017simple} to associate detections across frames and construct identity-consistent vehicle tracklets.
These tracklets provide temporally coherent centroid trajectories, which serve as mic-aligned spatial priors for associating globally mixed multi-channel audio with individual vehicle instances.

Since DeepSORT association in our implementation starts from the second frame, we apply first-frame completion (FC) to recover the first-frame box for each tracklet via greedy IoU matching with detections in the first frame.
We also use greedy IoU matching at the final frame for label inheritance, where each tracklet inherits the ground-truth vehicle-status label of the best-matched final-frame box.

\noindent\textbf{Tracklet-Conditioned Vehicle Status Classification.}
We propose the \textbf{TAVR classifier}, shown in \cref{fig:tavr_classifier}, a lightweight tracklet-level classifier that predicts the vehicle status at the final frame. 
Given a vehicle tracklet, the classifier uses three complementary inputs: the synchronized multi-channel audio spectrogram $A \in \mathbb{R}^{M \times T \times F}$, the absolute bounding-box centroid trajectory $\mathbf{BBoxC}$, and the centroid displacement trajectory $\boldsymbol{\Delta}\mathbf{BBoxC}$. 
The displacement trajectory provides a strong motion cue for separating moving vehicles from stationary ones, while the absolute centroid trajectory provides a geometry-conditioned spatial cue for associating the globally mixed multi-channel audio with the corresponding vehicle instance. 
We empirically validate the importance of these input components in \cref{sec:tavr_ablations}.

Specifically, we define the bounding-box centroid at time $t$ as
$\mathbf{BBoxC}^{t}=(c_x^t,c_y^t) \in \mathbb{R}^{2}$.
The centroid displacement between consecutive frames is then computed as:
\begin{equation}
\label{eq:displacement}
\begin{aligned}
\boldsymbol{\Delta}\mathbf{BBoxC}^{t} &= \mathbf{BBoxC}^{t+1} - \mathbf{BBoxC}^{t}, \\
t &= 0,\ldots,L-2.
\end{aligned}
\end{equation}

The classifier is formulated as:
\begin{equation}
\label{eq:tavr_classifier}
\mathrm{cls}=f_{\text{TAVR}}(\boldsymbol{\Delta}\mathbf{BBoxC}, \mathbf{BBoxC}, A).
\end{equation}

As shown in \cref{fig:tavr_classifier}, the absolute centroid positions $\mathbf{BBoxC}$ and relative centroid displacements $\boldsymbol{\Delta}\mathbf{BBoxC}$ are first embedded and concatenated to form vehicle-conditioned query tokens. 
The multi-channel audio spectrogram $A$ is encoded into cross-channel audio tokens, which serve as keys and values. 
Through cross-attention, the classifier retrieves audio evidence conditioned on each vehicle's tracklet geometry, enabling instance-specific audio attribution from the globally mixed audio input. 
The retrieved audio-geometry features are then reshaped along the temporal dimension and passed to a bidirectional GRU to aggregate tracklet-level temporal information. 
Finally, the temporally aggregated representation is fed into a classification layer to predict the final-frame vehicle status.

\noindent\textbf{Joint Audio-visual Contrastive Embedding (JACE).}
To improve class-wise separation and cross-domain robustness, we regularize the audio-geometry representation with a supervised contrastive objective~\cite{khosla2020supervised}.
Specifically, for each vehicle tracklet $i$, we concatenate its absolute bounding-box centroid position $\mathbf{BBoxC}_i$, centroid displacement $\boldsymbol{\Delta}\mathbf{BBoxC}_i$, and audio feature $A_i$, and project the concatenated feature into a joint latent embedding:
\begin{equation}
    \label{eq:joint_concatenation}
    \mathbf{c}_i = \operatorname{proj}\big(
        \operatorname{cat}(
            \mathbf{BBoxC}_i,\,
            \boldsymbol{\Delta}\mathbf{BBoxC}_i,\,
            A_i
        )
    \big),
\end{equation}

\begin{equation}
    \mathcal{L}_{\text{sup}} 
    = \sum_{i \in \mathcal{I}} 
    \frac{-1}{|P(i)|}
    \sum_{p \in P(i)}
    \log 
    \frac{
        \exp\left(\frac{\mathbf{c}_i^\top \mathbf{c}_p}{\tau}\right)
    }{
        \sum\limits_{a \in \mathcal{I} \setminus \{i\}}
        \exp\left(\frac{\mathbf{c}_i^\top \mathbf{c}_a}{\tau}\right)
    },
\end{equation}

where $P(i)$ denotes the set of tracklets sharing the same vehicle-status label as tracklet $i$, and $\tau$ is the temperature parameter.
This supervised contrastive loss pulls tracklets with the same status label closer while pushing tracklets from different status classes apart in the joint audio-geometry space.

We supervise training in a multi-task manner. The overall objective is the sum of a classification loss and the supervised contrastive loss:
\begin{equation}
    \mathcal{L} = \mathcal{L}_{\text{sup}} + \mathcal{L}_{\text{ce}}.
\end{equation}
This JACE objective constructs a task-specific representation space defined by audio-geometry correspondence: Moving corresponds to a moving tracklet with high-RPM vehicle sound, Idling corresponds to a static tracklet with low-RPM engine sound, and Engine-Off corresponds to a static tracklet without engine sound.
During inference, we fuse the TAVR classifier predictions with the associated bounding boxes to obtain the final predictions.

\subsection{Training-Free Cross-Domain Channel Alignment}
\label{sec:channel_alignment}
In practical roadside surveillance deployments, sensor configurations can vary across installation sites. A common deployment variation is the inadvertent reordering of microphone connections by field technicians, which changes the input channel ordering. Since the learned audio-geometry correspondence depends on a consistent physical microphone layout, an incorrect channel order can break the geometry-conditioned binding between vehicle tracklets and multi-channel audio.

To address this deployment challenge without retraining the TAVR classifier, we use a lightweight training-free channel-correspondence alignment procedure. Given a small labeled calibration set from the target deployment site, we evaluate candidate microphone-channel correspondence mappings while keeping all model parameters frozen. The selected mapping is the one that gives the best calibration performance under the frozen TAVR model. This procedure is label-assisted but training-free: target-domain labels are used only to select the channel correspondence, not to update the detector, the TAVR classifier, or JACE.

After the correspondence mapping is selected, the reordered audio is used for target-domain inference. This inference-time calibration introduces negligible computational overhead and allows the learned audio-geometry representation to be transferred to deployment sites with different microphone channel orderings.
\section{Experiments}
We evaluate TAVR in terms of detection accuracy, classifier robustness, and cross-domain generalization. We further analyze the quality of MASP tracklet construction and ablate the TAVR classifier components.

\noindent\textbf{Evaluation Metrics.} We use mAP$^{\mathrm{AD}}$ to denote the mean Average Precision for action detection (bbox + cls) and mAP$^{\mathrm{OD}}$ for object detection. For classification performance, we report the F1 score, mean Average Precision (mAP$^{\mathrm{CLS}}$), and Average Precision (AP$^{\mathrm{CLS}}$) to evaluate the TAVR classifier. For each predicted tracklet, we match its bounding box at each valid frame to the ground-truth box with the highest IoU in the same frame. We report five tracklet-level metrics following Table~\ref{tab:mot_quality}: \TMImean{} and \TMImed{} summarize the mean IoU over valid frames, \TMinImean{} summarizes the minimum IoU over time, \TIoU{} measures the fraction of valid frames with IoU above $0.5$, and \TCov{} measures the ratio of valid frames to the clip length. All metrics are averaged over tracklets.
\\
\noindent\textbf{Dataset.} We conduct experiments on the AVIVD dataset \cite{Li2024JointAI}. Following the original AVIVD data collection setup, we collect and annotate two additional test-only extensions: a same-location, different-day test set for evaluating temporal generalization, and a cross-location test set from a different deployment site for evaluating spatial and domain generalization. For clarity, we denote the original training set as AVIVD-LTrain, the original validation set as AVIVD-LV, the same-location different-day test set as AVIVD-LT, and the cross-location test set as AVIVD-M. AVIVD-LTrain contains 76,490 samples, with 26,924 moving, 36,968 idling, and 41,868 engine-off bounding boxes. AVIVD-LV contains 8,431 pairs, with 2,908 moving, 2,669 idling, and 3,422 engine-off bounding boxes. AVIVD-LT contains 2,380 pairs, with 78 moving, 785 idling, and 641 engine-off bounding boxes. AVIVD-M contains 177 pairs, with 24 moving, 166 idling, and 43 engine-off bounding boxes. \\
\noindent\textbf{Implementation Details.}  We implement all models using PyTorch and conduct training on RTX TITAN and RTX A6000 GPUs. More training details are included in supplementary materials. We will release the code upon acceptance.

\subsection{Comparison with Prior Arts}
We compare our method with three groups of baselines.
First, we include prior task-specific IVD methods, including Real-Time IVD~\cite{Li2023RealTimeIV}, AVIVDNet~\cite{Li2024JointAI}, and HAVT-IVD~\cite{Li2025HAVTIVDHC}, which represent existing two-stage and end-to-end solutions for roadside idling vehicle detection.
Second, we consider generic multimodal representation learning baselines, including ImageBind~\cite{girdhar2023imagebind}, LanguageBind~\cite{zhu2023languagebind}, AV-HuBERT~\cite{shi2022learning}, and CAV-MAE~\cite{gong2023contrastive}.
These methods are not designed specifically for IVD, but provide strong pretrained audio-visual or cross-modal representations and therefore serve as important transfer baselines.
Third, we include general-purpose audio-video MLLM baselines, MiniCPM-o~\cite{minicpmo2025} and VideoLLaMA 2~\cite{damonlpsg2024videollama2}, to examine whether large multimodal models with broad audio-video understanding ability can perform per-vehicle status reasoning. For generic representation and MLLM baselines, we use the same detector+MOT pipeline as TAVR to ensure identical instance anchors.
While TAVR represents each tracklet by per-frame bounding-box geometry and synchronized global audio, these baselines use the same per-frame boxes to crop vehicle regions from the original frames, forming an instance-level visual tracklet paired with the same global audio representation.
For generic representation baselines, we feed the cropped visual tracklet and audio representation into each pretrained model and train a lightweight downstream classifier on the AVIVD training split to predict the three IVD labels.
For MLLM baselines, we provide the cropped visual tracklet, a global multi-channel audio-feature visualization, and a fixed prompt describing the IVD classification rules; the model directly outputs a textual prediction, which is mapped to one of the three labels.
The exact prompt is provided in the supplementary material.
All methods are evaluated under the same final-frame per-vehicle detection protocol, enabling a unified comparison between prior IVD systems, generic multimodal representations, and audio-video MLLMs.

\begin{table*}[hbtp]
  \centering
  % \scriptsize
  \resizebox{\linewidth}{!}{
  \begin{tabular}{@{}l | c | c |
                  c c c c
                  c c c c |
                  c c c c
                  c c c c@{}}
    \toprule
    \multirow{2}{*}{\textbf{Method}} & \multirow{2}{*}{\textbf{Params}} & \multirow{2}{*}{\textbf{No Manual Mic Loc.}}
      & \multicolumn{4}{c}{\textbf{AVIVD-LV}}
      & \multicolumn{4}{c}{\textbf{AVIVD-LT}}
      & \multicolumn{4}{c}{\textbf{AVIVD-M (6 Mics)}}
      & \multicolumn{4}{c}{\textbf{AVIVD-M (3 Mics)}} \\
    \cmidrule(lr){4-7} \cmidrule(lr){8-11} \cmidrule(lr){12-15} \cmidrule(lr){16-19}
      &  &  & \textbf{mAP$^{\mathrm{AD}}$}
      & \textbf{AP$_{\mathrm{M}}^{\mathrm{AD}}$}
      & \textbf{AP$_{\mathrm{I}}^{\mathrm{AD}}$}
      & \textbf{AP$_{\mathrm{Eoff}}^{\mathrm{AD}}$}
      & \textbf{mAP$^{\mathrm{AD}}$}
      & \textbf{AP$_{\mathrm{M}}^{\mathrm{AD}}$}
      & \textbf{AP$_{\mathrm{I}}^{\mathrm{AD}}$}
      & \textbf{AP$_{\mathrm{Eoff}}^{\mathrm{AD}}$}
      & \textbf{mAP$^{\mathrm{AD}}$}
      & \textbf{AP$_{\mathrm{M}}^{\mathrm{AD}}$}
      & \textbf{AP$_{\mathrm{I}}^{\mathrm{AD}}$}
      & \textbf{AP$_{\mathrm{Eoff}}^{\mathrm{AD}}$}
      & \textbf{mAP$^{\mathrm{AD}}$}
      & \textbf{AP$_{\mathrm{M}}^{\mathrm{AD}}$}
      & \textbf{AP$_{\mathrm{I}}^{\mathrm{AD}}$}
      & \textbf{AP$_{\mathrm{Eoff}}^{\mathrm{AD}}$} \\
    \midrule
      &  &  &  \multicolumn{8}{c}{\makecell{\textbf{Source Domain}\\{\scriptsize P: 2-3-0-1-4-5, S: 2.4m}}}
      & \multicolumn{4}{c}{\makecell{\textbf{Target Domain}\\{\scriptsize P: 3-0-2-4-1-5, S: 2.4m}}}
      &  \multicolumn{4}{c}{\makecell{\textbf{Target Domain}\\{\scriptsize P: 3-2-5, S: 4.8m}}} \\
    \midrule
    \rowcolor{gray!15}
    \multicolumn{19}{@{}l}{\textbf{Task-Specific Models}} \\
    Real-Time IVD \cite{Li2023RealTimeIV} & 2.40M & \xmark
      & $80.97$ & $92.45$ & $68.93$ & $81.55$
      & $59.20$ & $70.40$ & $40.23$ & $66.98$
      & $1.39$ & $4.17$ & $0.00$ & $0.00$
      & $1.39$ & $4.17$ & $0.00$ & $0.00$ \\
    AVIVDNet \cite{Li2024JointAI} & 4.33M & \cmark
      & $79.21$ & $\underline{93.43}$ & $66.74$ & $77.47$
      & $30.83$ & $75.39$ & $0.23$ & $16.87$
      & $0.00$ & $0.00$ & $0.00$ & $0.00$
      & $0.00$ & $0.00$ & $0.00$ & $0.00$  \\
    HAVT-IVD (ICASSP'26) \cite{Li2025HAVTIVDHC} & 31.62M & \cmark
      & $\underline{88.63}$ & $\textbf{94.35}$ & $83.41$ & $88.12$
      & $43.34$ & $\textbf{91.83}$ & $38.20$ & $0.00$
      & $0.00$ & $0.00$ & $0.00$ & $0.00$
      & $0.00$ & $0.00$ & $0.00$ & $0.00$  \\
    \midrule
    \rowcolor{gray!15}
    \multicolumn{19}{@{}l}{\textbf{Generic Audio-visual Representation Learning Baselines / MLLMs}} \\
    ImageBind \cite{girdhar2023imagebind} & 721.8M & \cmark
      & $76.17$ & $74.69$ & $65.44$ & $\underline{88.38}$
      & $\underline{66.10}$ & $63.46$ & $45.60$ & $\mathbf{89.24}$
      & $\underline{46.07}$ & $\underline{52.38}$ & $75.84$ & $\underline{9.99}$ 
      & $34.23$ & $\underline{73.71}$ & $1.98$ & $27.01$ \\
    LanguageBind \cite{zhu2023languagebind} & 711.7M & \cmark
      & $73.56$ & $75.96$ & $\underline{73.65}$ & $71.08$
      & $37.30$ & $68.29$ & $25.45$ & $18.17$
      & $34.80$ & $24.54$ & $\underline{79.87}$ & $0.00$
      & $43.87$ & $54.46$ & $\mathbf{75.67}$ & $1.49$ \\
    AV-HuBERT \cite{shi2022learning} & 102.95M & \cmark
      & $44.11$ & $40.50$ & $41.51$ & $50.32$
      & $40.95$ & $68.50$ & $\underline{54.24}$ & $0.12$
      & $11.93$ & $33.98$ & $1.81$ & $0.00$
      & $\underline{50.98}$ & $64.80$ & $42.57$ & $\underline{45.56}$ \\
    CAV-MAE \cite{gong2023contrastive} & 166.9M & \cmark
      & $51.91$ & $59.59$ & $47.21$ & $48.92$
      & $16.66$ & $31.53$ & $18.46$ & $0.00$
      & $15.20$ & $10.02$ & $35.57$ & $0.00$
      & $3.43$ & $10.30$ & $0.00$ & $0.00$ \\
    MiniCPM-o \cite{minicpmo2025} & 8.7B & \cmark
      & $8.58$ & $25.56$ & $0.19$ & $0.00$
      & $19.45$ & $6.83$ & $1.03$ & $0.00$
      & $5.84$ & $17.53$ & $0.00$ & $0.00$
      & $5.84$ & $17.53$ & $0.00$ & $0.00$  \\
    VideoLLaMA 2 \cite{damonlpsg2024videollama2} & 8.42B & \cmark
      & $8.64$ & $25.49$ & $0.29$ & $0.15$
      & $6.54$ & $19.35$ & $0.26$ & $0.00$
      & $6.16$ & $14.36$ & $4.14$ & $0.00$
      & $14.36$ & $15.06$ & $28.01$ & $0.00$ \\
    % Qwen3.5-Omni-Plus \cite{xu2025qwen3} & Undisclosed & \cmark
    %   &  &  &  & 
    %   &  &  &  & 
    %   &  &  &  & 
    %   &  &  &  &  \\ 
    \midrule
    TAVR (\textbf{Ours}) & 8.89M & \cmark
      & $\mathbf{89.59}$ & $90.43$ & $\mathbf{88.33}$ & $\mathbf{90.00}$
      & $\mathbf{81.32}$ & $\underline{91.02}$ & $\mathbf{69.10}$ & $\underline{83.84}$
      & $\mathbf{78.10}$ & $\mathbf{82.61}$ & $\mathbf{86.14}$ & $\mathbf{65.55}$
      & $\mathbf{67.90}$ & $\mathbf{83.17}$ & $\underline{69.19}$ & $\mathbf{51.35}$ \\
    \bottomrule
  \end{tabular}
  }
  \caption{Comparison with task-specific IVD methods, generic audio-visual representation baselines, and audio-video MLLMs across same-day (AVIVD-LV), cross-day (AVIVD-LT) and cross-domain (AVIVD-M (6 Mics, 3 Mics)) settings. No Manual Mic Loc. indicates whether the method can be deployed without manually annotating microphone pixel locations in the camera frame. Channel order denotes microphone indices from left to right, and spacing denotes inter-microphone distance.}
  \label{tab:comparison_ivd}
\end{table*}

As shown in Table~\ref{tab:comparison_ivd}, AVIVD-LV evaluates
same-day source-domain performance. TAVR achieves the best overall $\mathrm{mAP}^{\mathrm{AD}}$ of 89.59, slightly outperforming HAVT-IVD~\cite{Li2025HAVTIVDHC} (88.63) while using far fewer parameters (8.89M vs. 31.62M), suggesting that the gain comes from the trackletized formulation rather than increased model scale. The gain mainly comes from more
balanced class-wise detection: TAVR obtains the best AP for Idling
($88.33$) and Engine-Off ($90.00$), whereas HAVT-IVD~
\cite{Li2025HAVTIVDHC} is strongest on Moving ($94.35$) but
lower on the two stationary classes. Compared with AVIVDNet~
\cite{Li2024JointAI} and Real-Time IVD~\cite{Li2023RealTimeIV},
TAVR improves mAP$^{\mathrm{AD}}$ by $10.38$ and $8.62$ points,
respectively. Among generic representation models, ImageBind~
\cite{girdhar2023imagebind} and LanguageBind~\cite{zhu2023languagebind}
show competitive single-class behavior, e.g., ImageBind~\cite{girdhar2023imagebind}
reaches $88.38$ AP on Engine-Off and LanguageBind~\cite{zhu2023languagebind}
reaches $73.65$ AP on Idling. However, their lower mAP$^{\mathrm{AD}}$
indicates that generic cross-modal representations do not consistently
separate all three vehicle states. The MLLM baselines MiniCPM-o~
\cite{minicpmo2025} and VideoLLaMA 2~\cite{damonlpsg2024videollama2}
perform poorly on AVIVD-LV, suggesting that broad audio-video
understanding alone is insufficient for object-level roadside IVD.

AVIVD-LT further tests cross-day generalization at the same deployment
site. TAVR remains the best method with $81.32$ mAP$^{\mathrm{AD}}$,
exceeding the strongest baseline ImageBind~\cite{girdhar2023imagebind}
($66.10$) by $15.22$ points. This split exposes a clear robustness gap
among task-specific models: Real-Time IVD~\cite{Li2023RealTimeIV}
drops from $80.97$ to $59.20$, AVIVDNet~\cite{Li2024JointAI}
drops from $79.21$ to $30.83$, and HAVT-IVD~\cite{Li2025HAVTIVDHC}
drops from $88.63$ to $43.34$, while TAVR only drops from $89.59$ to
$81.32$. Although HAVT-IVD~\cite{Li2025HAVTIVDHC} keeps the
highest Moving AP ($91.83$), it fails on Engine-Off ($0.00$), and
AVIVDNet~\cite{Li2024JointAI} nearly collapses on Idling
($0.23$). ImageBind~\cite{girdhar2023imagebind} obtains the
best Engine-Off AP ($89.24$), but its Idling AP is only $45.60$.
In contrast, TAVR maintains strong performance across all classes,
with $91.02$ AP for Moving, $69.10$ for Idling, and $83.84$ for
Engine-Off. These results show that the proposed tracklet-conditioned
audio-geometry reasoning improves not only same-day accuracy, but also
cross-day stability, especially for the two stationary states that
require fine-grained audio-spatial association rather than motion cues
alone.

%%%%%%%%%%%% 放在附录
\subsection{MASP Construction Performance}
MASP relies on reliable vehicle anchors before tracklet-level audio-visual reasoning. In \cref{tab:od_mot_performance}, we therefore evaluate its construction quality from two perspectives: frame-level object detection and tracklet-level quality after DeepSORT tracking~\cite{wojke2017simple}. The former evaluates whether vehicles can be accurately detected and localized in each individual frame, while the latter evaluates whether DeepSORT can link these frame-wise detections into temporally continuous and spatially stable vehicle tracklets for downstream status classification. Since roadside IVD follows a task-specific annotation protocol that excludes vehicles outside the target area, we fine-tune YOLOv11 detectors on the AVIVD training split and a small set of $942$ AVIVD-M images with bounding-box annotations only. These $942$ images provide target-domain detector calibration only: they do not include vehicle-status labels, and no AVIVD-M status labels are used to train TAVR classifier with JACE. We distinguish detector calibration from vehicle-status adaptation: the AVIVD-M bounding boxes are used only to obtain reliable target-site vehicle anchors, while all AVIVD-M vehicle-status labels remain unseen during TAVR classifier and JACE training.

\begin{table*}[htbp]
  \centering
  \caption{Object detection and DeepSORT-based tracklet bounding-box quality on AVIVD-LV, AVIVD-LT, and AVIVD-M.}
  \label{tab:od_mot_performance}

  \begin{minipage}[t]{0.58\textwidth}
    \centering
    \textbf{(a) Object detection performance.}
    \phantomsubcaption\label{tab:od_performance}

    \vspace{0.5em}

    \resizebox{\textwidth}{!}{
    \begin{tabular}{lccccccccc}
      \toprule
      \multirow{2}{*}{Detector} 
        & \multicolumn{3}{c}{\textbf{AVIVD-LV}} 
        & \multicolumn{3}{c}{\textbf{AVIVD-LT}}
        & \multicolumn{3}{c}{\textbf{AVIVD-M}} \\
      \cline{2-10}
        & $mAP@0.5$ & $mAP@0.75$ & $mAP@avg$ 
        & $mAP@0.5$ & $mAP@0.75$ & $mAP@avg$
        & $mAP@0.5$ & $mAP@0.75$ & $mAP@avg$ \\
      \midrule
      YOLOv11n & $99.00$ & $87.23$ & $75.35$ & $99.01$ & $16.87$ & $42.74$ & $99.01$ & $97.81$ & $85.08$ \\
      YOLOv11s & $99.00$ & $86.47$ & $77.14$ & $99.01$ & $18.70$ & $43.60$ & $100.00$ & $99.01$ & $83.69$ \\
      YOLOv11m & $98.95$ & $95.10$ & $76.05$ & $98.02$ & $68.88$ & $65.83$ & $99.01$ & $82.13$ & $81.73$ \\
      YOLOv11l & $99.00$ & $85.18$ & $76.53$ & $99.01$ & $21.71$ & $49.90$ & $99.01$ & $80.35$ & $82.51$ \\
      YOLOv11x & $98.99$ & $88.02$ & $77.49$ & $99.01$ & $19.15$ & $42.85$ & $99.01$ & $81.92$ & $79.20$ \\
      \bottomrule
    \end{tabular}
    }
  \end{minipage}
  \hfill
  \begin{minipage}[t]{0.38\textwidth}
    \centering
    \textbf{(b) Tracklet-level bounding-box quality of DeepSORT~\cite{wojke2017simple} outputs.}
    \phantomsubcaption\label{tab:mot_quality}

    \vspace{0.5em}

    \resizebox{\textwidth}{!}{
    \begin{tabular}{llccccc}
      \toprule
      Dataset & Detector+DeepSORT 
        & \TMImean 
        & \TMImed 
        & \TMinImean 
        & \TIoU 
        & \TCov \\
      \midrule
      AVIVD-LV & YOLOv11s + DeepSORT 
        & $83.18$ & $85.63$ & $80.09$ & $98.75$ & $89.60$ \\
      AVIVD-LT & YOLOv11s + DeepSORT
        & $70.55$ & $64.40$ & $69.30$ & $99.66$ & $92.37$ \\
      AVIVD-M & YOLOv11s + DeepSORT
        & $89.97$ & $92.08$ & $88.85$ & $100.00$ & $93.40$ \\
      \bottomrule
    \end{tabular}
    }
  \end{minipage}
\end{table*}

As shown in \cref{tab:od_performance}, all YOLOv11 variants achieve nearly saturated $mAP@0.5$ on AVIVD-LV, AVIVD-LT, and AVIVD-M, indicating that vehicle recall is not the main bottleneck for MASP construction. The main difference appears at stricter localization thresholds. In particular, AVIVD-LT has much lower $mAP@0.75$ and $mAP@avg$ than AVIVD-LV, even though its $mAP@0.5$ remains around $99\%$. We attribute this gap mainly to annotation-style variation across different annotators and collection days: some annotators draw vehicle boxes slightly tighter, whereas others include a looser extent of the vehicle. Such small boundary differences are heavily penalized by high-IoU metrics, but they do not imply that the detector fails to localize the vehicle. This localization variation has limited impact on TAVR classifier because it relies on centroid and displacement rather than dense box shape or exact box boundaries. Consistent with this design, \cref{tab:mot_quality} shows that YOLOv11s+DeepSORT~\cite{wojke2017simple} produces reliable tracklets across all splits, with high \TIoU{} and \TCov{} on AVIVD-LV, AVIVD-LT, and AVIVD-M. Although AVIVD-LT has lower tracklet-box overlap scores, its temporal coverage remains strong, suggesting that MASP provides sufficiently stable instance anchors for the subsequent audio-geometry status reasoning stage.

%%%%%%%%%%%%
\begin{table*}[htbp]
  \centering
  \resizebox{\linewidth}{!}{
  \begin{tabular}{@{}ccccc c ccccc ccccc ccccc@{}}
    \toprule
    \multicolumn{5}{c}{\textbf{Input Components}} 
      & \multirow{2}{*}{\textbf{Input Source}}
      & \multicolumn{5}{c}{\textbf{AVIVD-LV}} 
      & \multicolumn{5}{c}{\textbf{AVIVD-LT}} 
      & \multicolumn{5}{c}{\textbf{AVIVD-M}} \\
    \cmidrule(lr){1-5} \cmidrule(lr){7-11} \cmidrule(lr){12-16} \cmidrule(lr){17-21}
    \textbf{A} & \textbf{BBox} & \textbf{Dis} & \textbf{VC} & \textbf{JACE}
      &
      & \textbf{F1} & \textbf{mAP$^{\mathrm{CLS}}$} & AP(M) & AP(I) & AP(Eoff) 
      & \textbf{F1} & \textbf{mAP$^{\mathrm{CLS}}$} & AP(M) & AP(I) & AP(Eoff)
      & \textbf{F1} & \textbf{mAP$^{\mathrm{CLS}}$} & AP(M) & AP(I) & AP(Eoff) \\
    \midrule
    \checkmark &        & \checkmark &        &       
      & OD
      & $82.97$ & $75.81$ & $92.07$ & $62.88$ & $72.48$
      & $69.48$ & $64.91$ & $70.89$ & $67.05$ & $56.79$
      & $69.96$ & $63.70$ & $\textbf{84.81}$ & $80.50$ & $25.79$ \\
    \checkmark & \checkmark &        &        &       
      & OD
      & $85.07$ & $77.49$ & $80.68$ & $69.10$ & $82.69$
      & $28.10$ & $35.38$ & $11.02$ & $52.19$ & $42.92$
      & $28.20$ & $33.93$ & $10.30$ & $74.31$ & $17.17$ \\
    \checkmark & \checkmark & \checkmark &        &       
      & OD
      & $89.56$ & $83.88$ & $90.01$ & $77.15$ & $84.48$
      & $75.18$ & $65.87$ & $48.51$ & $76.48$ & $72.61$
      & $41.93$ & $57.82$ & $81.31$ & $71.76$ & $20.38$ \\
    \checkmark &        &        & \checkmark & \checkmark
      & OD 
      & $74.48$ & $65.91$ & $57.08$ & $63.50$ & $77.16$
      & $22.67$ & $33.87$ & $9.57$ & $49.20$ & $42.82$
      & $36.81$ & $39.43$ & $26.09$ & $73.76$ & $18.45$ \\
    \checkmark & \checkmark & \checkmark & \checkmark & \checkmark
      & OD 
      & $88.84$ & $83.09$ & $86.77$ & $75.41$ & $87.08$
      & $46.26$ & $50.58$ & $57.39$ & $51.53$ & $42.82$
      & $48.59$ & $53.95$ & $74.79$ & $68.60$ & $18.45$ \\
    \checkmark & \checkmark & \checkmark &        & \checkmark
      & OD
      & $\textbf{94.97}$ & $\textbf{91.87}$ & $\textbf{93.67}$ 
      & $\textbf{87.42}$ & $\textbf{94.54}$ 
      & $\textbf{80.03}$ & $\textbf{71.50}$ & $\textbf{66.81}$ 
      & $\textbf{76.67}$ & $\textbf{71.01}$
      & $\textbf{86.08}$ & $\textbf{79.81}$ & $81.08$ & $\textbf{96.02}$ & $\textbf{62.32}$ \\
    \hline
    \hline
    \checkmark & \checkmark & \checkmark &        & \checkmark
      & GT
      & $96.52$ & $94.35$ & $97.77$ 
      & $90.40$ & $94.88$ 
      & $78.65$ & $74.64$ & $84.97$ 
      & $76.23$ & $62.71$
      & $76.34$ & $73.77$ & $92.00$ & $88.74$ & $40.57$ \\
    \bottomrule
  \end{tabular}
  }
  \caption{
Ablation study of the TAVR classifier with different input combinations on AVIVD-LV, AVIVD-LT, and AVIVD-M.
A denotes 6-channel audio, BBox denotes the bounding-box centroid position, Dis denotes bounding-box displacement, and VC denotes vehicle crop.
Results are reported using F1, $\mathbf{mAP}^{\mathrm{CLS}}$, and class-wise AP for Moving (M), Idling (I), and Engine-off (Eoff).
OD indicates tracklets generated by the detector and MOT, while GT indicates tracklets constructed from ground-truth boxes and MOT.
}
  \label{tab:tavr_classifier_ablation}
\end{table*}

\begin{figure}[htbp]
  \centering
  \subfloat[AVIVD-LV with JACE.]{%
    \includegraphics[width=0.5\linewidth]{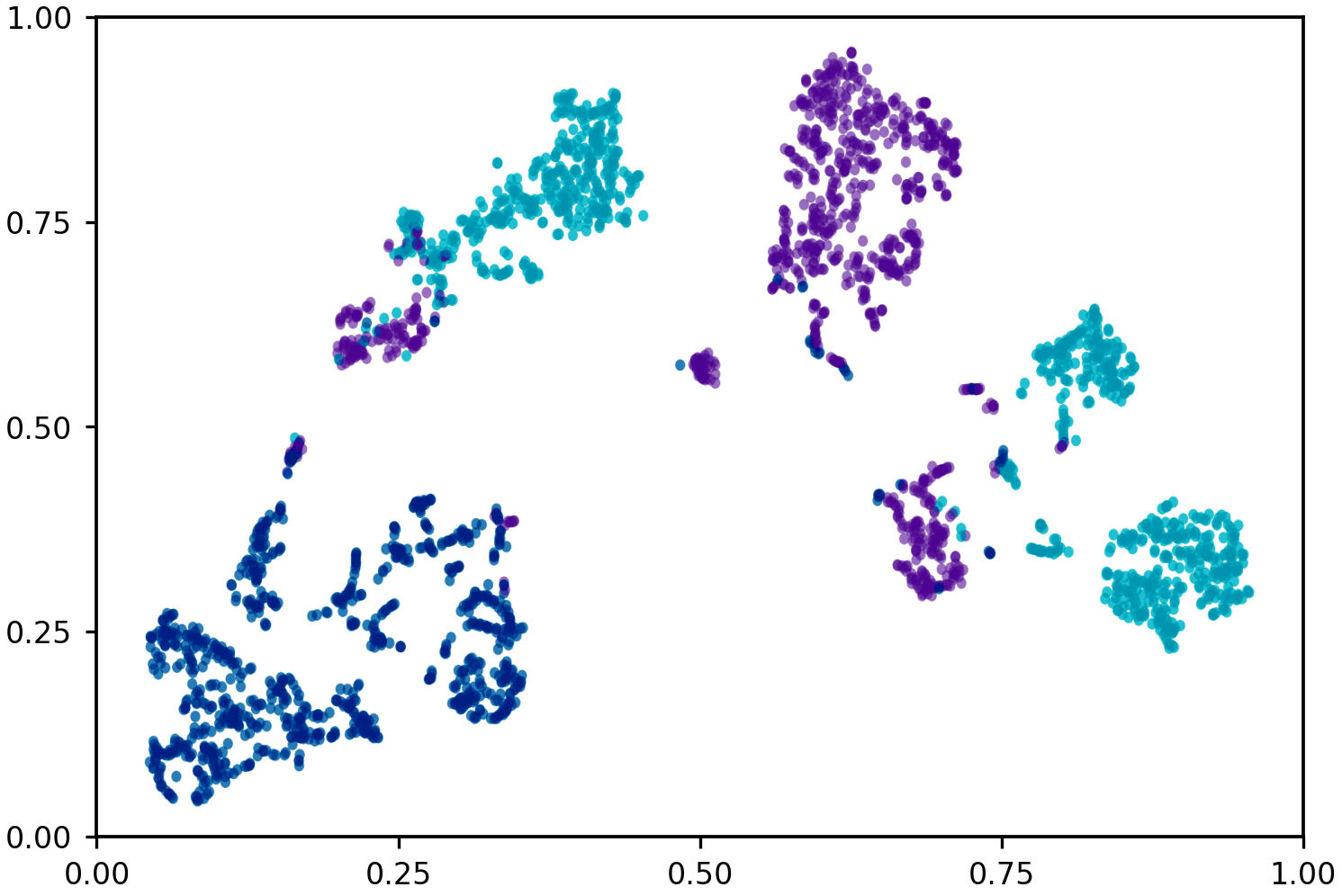}%
  }\hfill
  \subfloat[AVIVD-LV without JACE.]{%
    \includegraphics[width=0.5\linewidth]{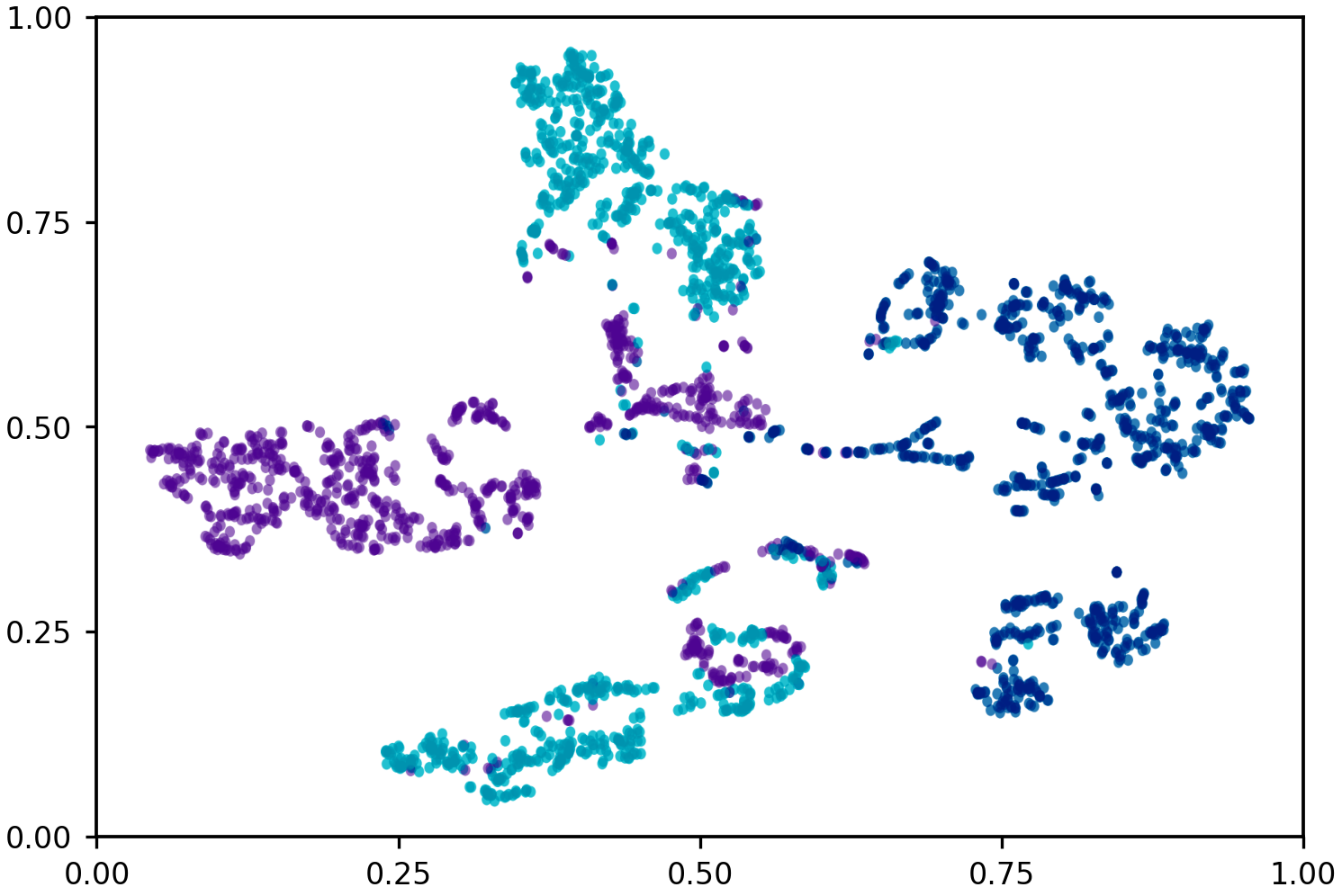}%
  }\\[0.5em]
  \subfloat[AVIVD-LT with JACE.]{%
    \includegraphics[width=0.5\linewidth]{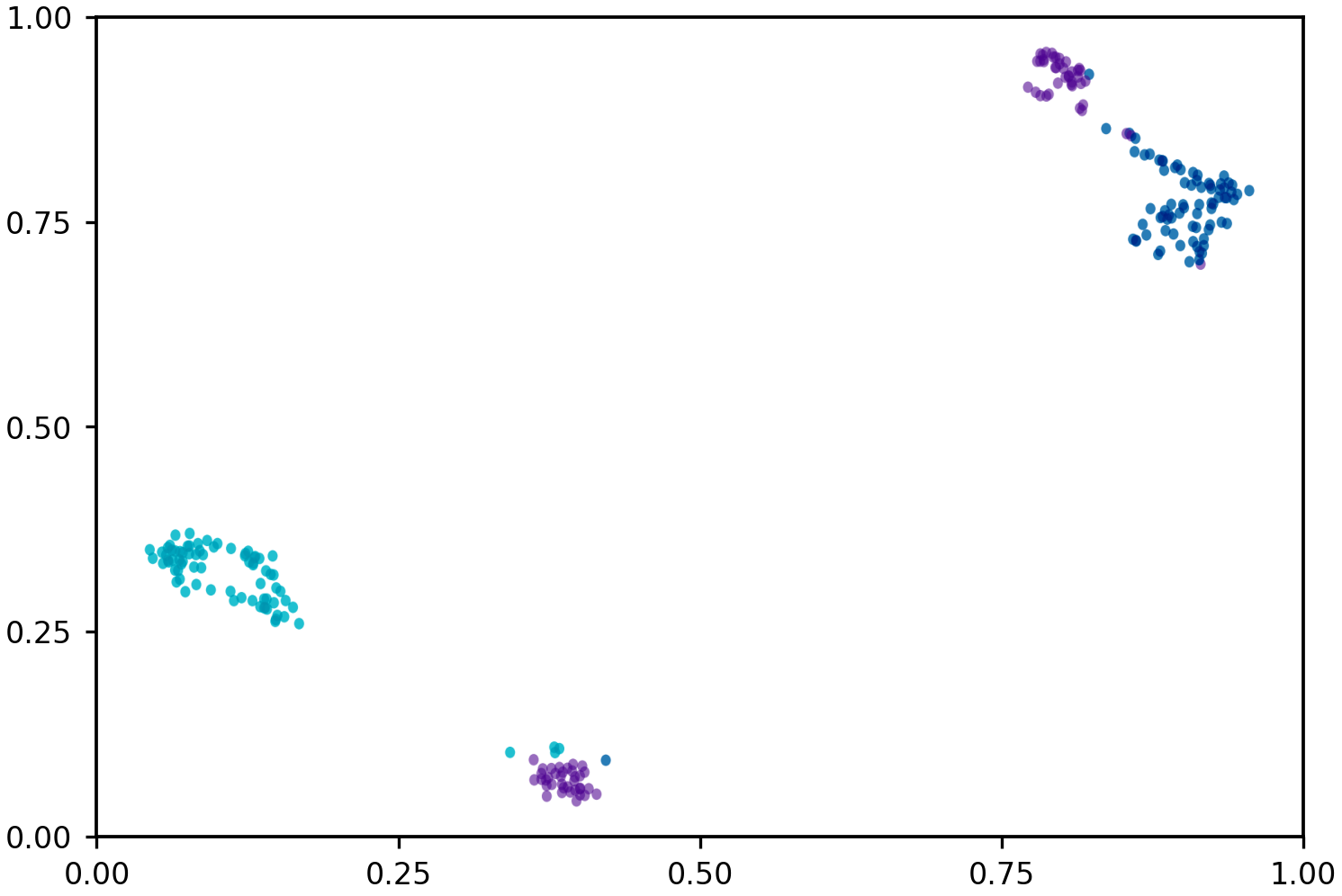}%
  }\hfill
  \subfloat[AVIVD-LT without JACE.]{%
    \includegraphics[width=0.5\linewidth]{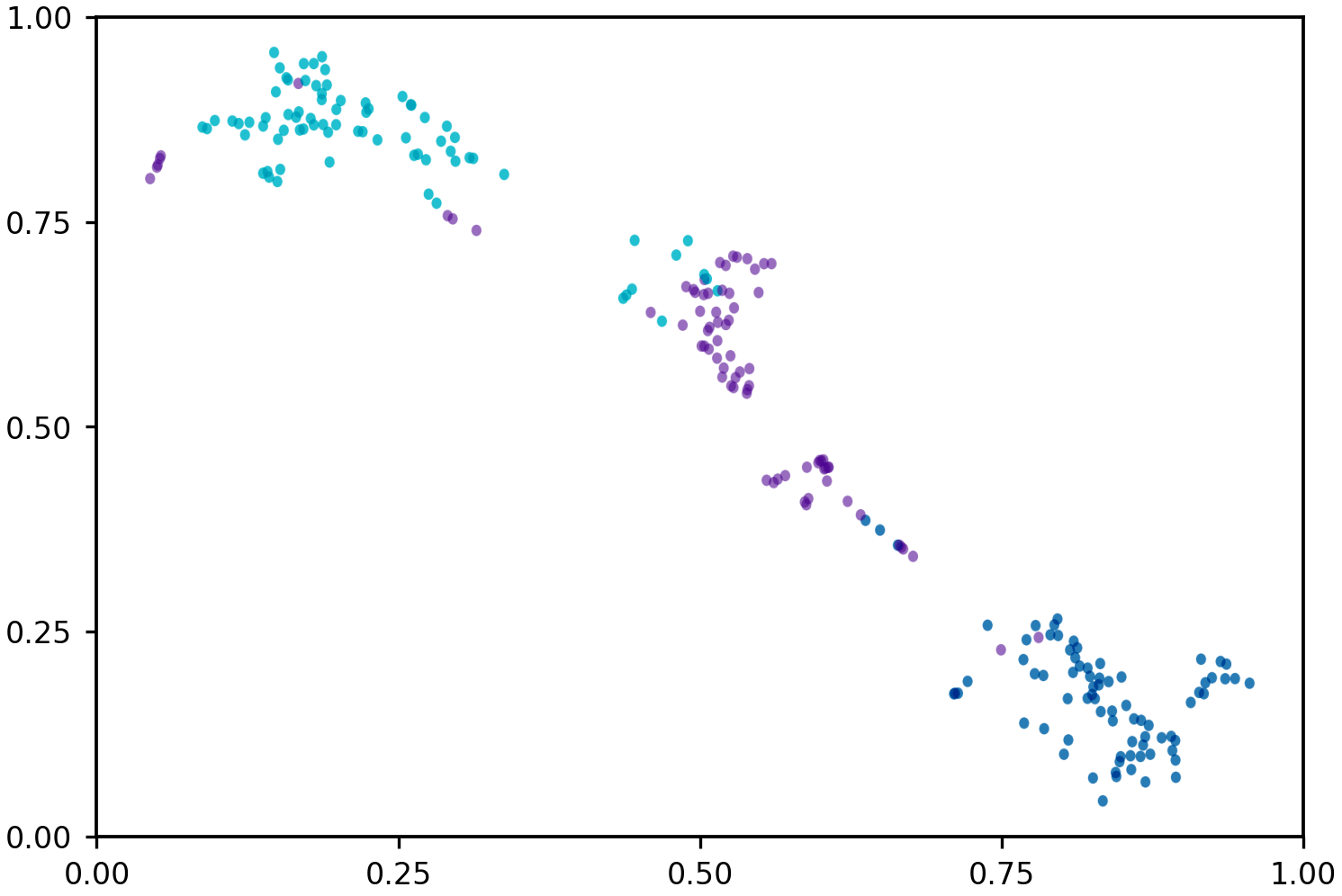}%
  }\\[0.5em]
  \subfloat[AVIVD-M with JACE.]{%
    \includegraphics[width=0.5\linewidth]{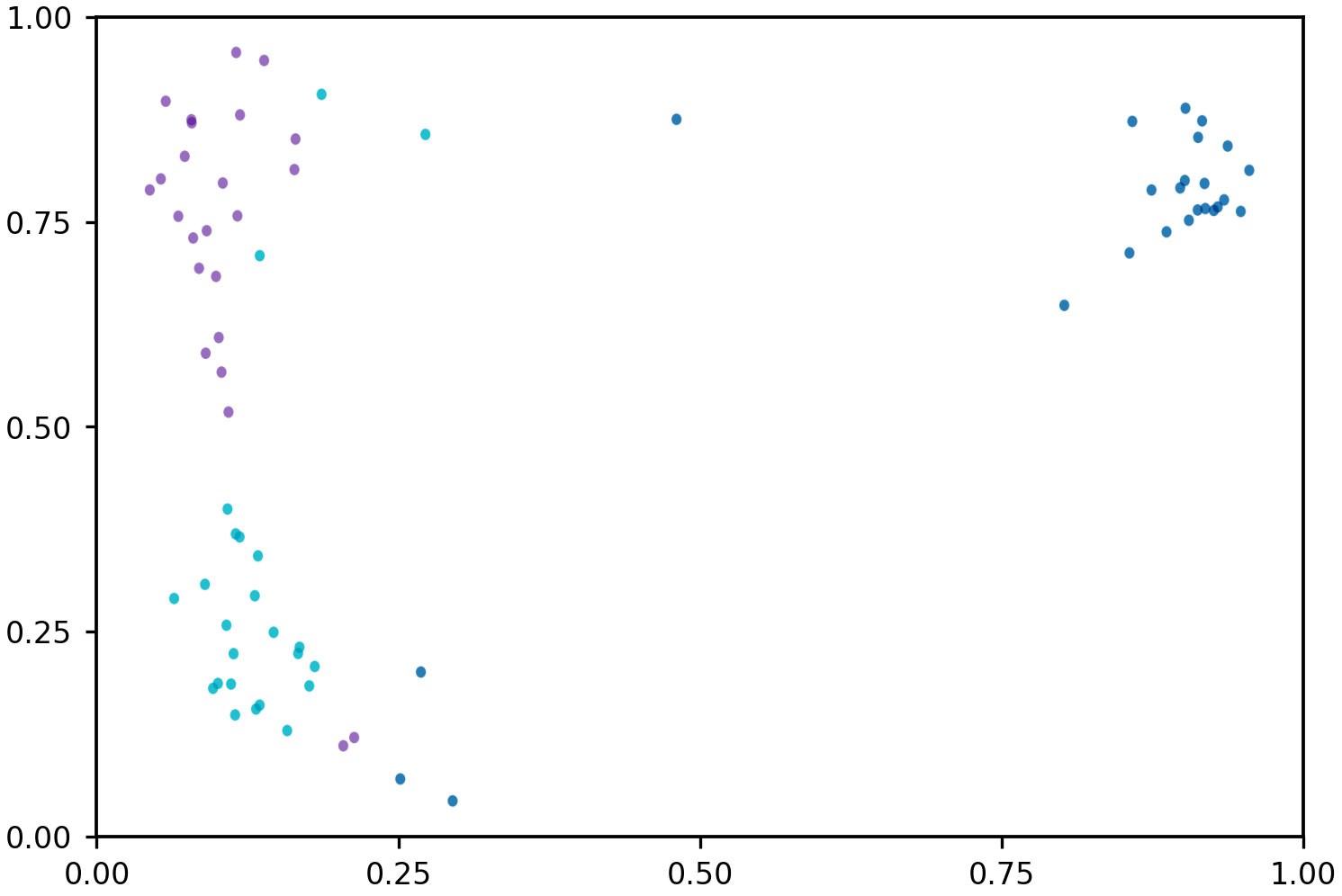}%
  }\hfill
  \subfloat[AVIVD-M without JACE.]{%
    \includegraphics[width=0.5\linewidth]{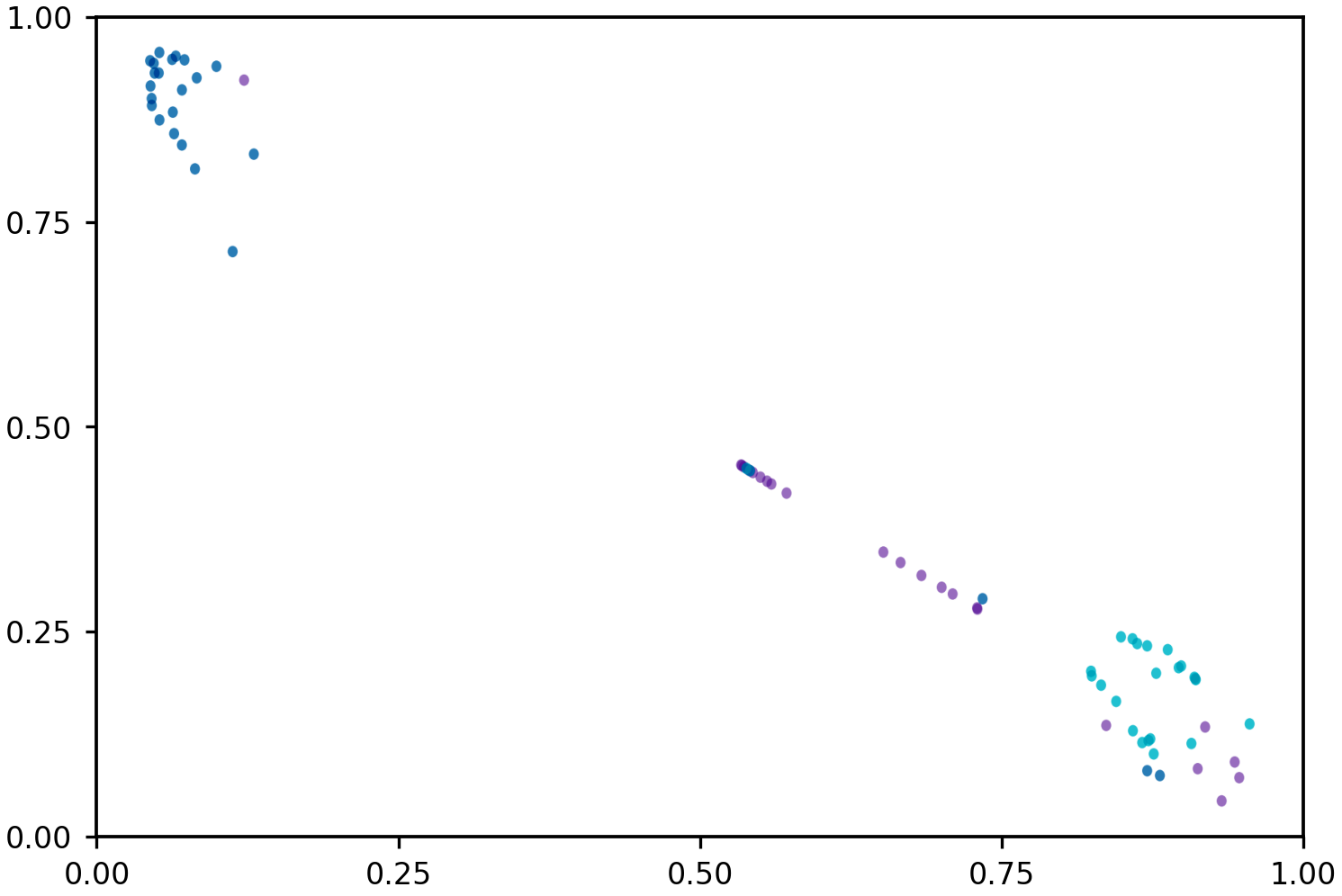}%
  }
  \caption{t-SNE visualization of TAVR classifier latent representations with and without JACE across three evaluation splits. In each row, the left panel uses JACE and the right panel removes JACE. Across all three splits, JACE yields more compact and better separated clusters for Moving, Idling, and Engine-Off states.}
  \label{fig:tSNE}
\end{figure}

\subsection{Ablations on TAVR Classifier}
\label{sec:tavr_ablations}
We evaluate the effectiveness of the TAVR classifier by
ablating its input components and the proposed JACE
regularization. Table~\ref{tab:tavr_classifier_ablation} reports
classification performance on AVIVD-LV, AVIVD-LT, and
AVIVD-M. We summarize four key observations. (1) Motion and spatial
geometry are complementary: $A+\mathrm{Dis}$ performs well on
AVIVD-LV and AVIVD-LT, but its Engine-Off AP on AVIVD-M is only
25.79, while $A+\mathrm{BBox}$ also drops strongly under domain shift.
Combining audio, absolute position, and displacement improves the
classifier to 83.88 on AVIVD-LV and 65.87 on AVIVD-LT, confirming the
need for both tracklet motion and geometry-conditioned audio binding.
(2) Visual crops are less reliable than tracklet geometry: the
$A+\mathrm{VC}+\mathrm{JACE}$ variant remains low on all splits, and
adding VC to $A+\mathrm{BBox}+\mathrm{Dis}+\mathrm{JACE}$ reduces mAP
from 71.50 to 50.58 on AVIVD-LT and from 79.81 to 53.95 on AVIVD-M.
This suggests that, unlike ASD face/lip crops, vehicle crops introduce appearance/background variation and provide weaker status cues than compact tracklet geometry. (3) JACE
consistently improves robustness: adding it to
$A+\mathrm{BBox}+\mathrm{Dis}$ increases mAP$_{\mathrm{CLS}}$ from
83.88 to 91.87 on AVIVD-LV, from 65.87 to 71.50 on AVIVD-LT, and from
57.82 to 79.81 on AVIVD-M. The gains are especially important for
stationary classes, where status recognition depends more on
audio-geometry correspondence than motion alone. (4) Detector-generated
tracklets are sufficient for downstream classification: GT tracklets
improve AVIVD-LV and AVIVD-LT slightly, but OD tracklets remain
competitive and even outperform GT on AVIVD-M (79.81 vs. 73.77). This
is consistent with Table~\ref{tab:mot_quality}, indicating that the
tracking pipeline provides reliable anchors for TAVR.

Finally, t-SNE plots in Fig.~\ref{fig:tSNE} qualitatively support the JACE ablation in
Table~\ref{tab:tavr_classifier_ablation}. The difference between using
JACE and removing JACE is most pronounced on AVIVD-M, matching the
largest numerical gap in Table~\ref{tab:tavr_classifier_ablation}:
mAP$_{\mathrm{CLS}}$ improves from 57.82 to 79.81, F1 from 41.93 to
86.08, Idling AP from 71.76 to 96.02, and Engine-Off AP from 20.38 to
62.32. Correspondingly, the AVIVD-M t-SNE plots show that JACE produces
clearer class-wise clusters and margins, while the non-JACE features are
more entangled; this directly highlights the effectiveness of JACE for
building a status-discriminative audio-geometry latent space under
cross-domain shift.

\subsection{Performance under Cross-Domain Channel and Layout Shifts}
We evaluate the \textbf{cross-domain generalization} of our trained model on a different deployment domain, \textbf{AVIVD-M}, under two practical roadside deployment scenarios: channel-order variation caused by microphone reordering, and microphone-layout variation caused by different array sparsity and inter-microphone spacing. Table~\ref{tab:comparison_ivd} reports these two target-domain settings as AVIVD-M (6 Mics) and AVIVD-M (3 Mics), respectively. In the 6-microphone setting, the channel order changes from the source-domain order $2$-$3$-$0$-$1$-$4$-$5$ to $3$-$0$-$2$-$4$-$1$-$5$ while keeping the same $2.4$ m spacing. TAVR achieves $78.10$ mAP$^{\mathrm{AD}}$, substantially outperforming the strongest generic baseline ImageBind~\cite{girdhar2023imagebind} ($46.07$) and all prior task-specific IVD models, which collapse to near-zero performance under this cross-domain setting. In the more challenging 3-microphone setting, the array becomes sparser ($P:3$-$2$-$5$, $S:4.8$m), and TAVR still obtains $67.90$ mAP$^{\mathrm{AD}}$, outperforming the best baseline AV-HuBERT~\cite{shi2022learning} ($50.98$). This confirms that sparse-array deployment is indeed more difficult, a trend also observed in our prior HAVT-IVD study~\cite{Li2025HAVTIVDHC}. The degradation mainly appears in the stationary classes: from 6 mics to 3 mics, TAVR's AP for Idling and Engine-Off decreases from $86.14$ to $69.19$ and from $65.55$ to $51.35$, respectively, while Moving remains stable ($82.61$ to $83.17$). These results suggest that moving vehicles can still be recognized from robust motion-related evidence, whereas idling and engine-off vehicles require more precise audio-spatial alignment and are therefore more sensitive to microphone sparsity and spacing changes. The channel-correspondence search is training-free but label-assisted: AVIVD-M labels are used only to select the microphone-channel correspondence on a calibration set, while all model weights remain frozen.
We provide the detailed search procedure, permutation space, and correspondence-mapping convention in the supplementary material.

\section{Conclusion}
In conclusion, we propose TAVR, a tracklet-centric framework for roadside IVD that anchors vehicles with MASP, performs per-vehicle audio-geometry reasoning with centroid position and displacement, and uses JACE to learn a status-aware latent space. 
We further introduce AVIVD-LT and AVIVD-M to evaluate inter-day and cross-location generalization under realistic microphone configuration shifts. 
Experiments show that TAVR consistently outperforms prior IVD methods, generic audio-visual models, and audio-video MLLMs across AVIVD-LV, AVIVD-LT, and AVIVD-M. 
These results show that robust roadside IVD requires more than source-domain full-frame accuracy: grounding predictions on identity-consistent tracklets and audio-geometry correspondence leads to a more deployment-oriented solution. 
A remaining limitation is that TAVR can still be affected by incomplete or truncated audio observations, especially when distinguishing idling from engine-off vehicles. 
Future work will explore more robust audio representations and uncertainty-aware audio-geometry fusion under imperfect roadside sensing conditions.
\vfill
{
    \small
    \bibliographystyle{ieeenat_fullname}
    \bibliography{main}
}

\end{document}